\documentclass{article}

\usepackage[nonatbib, final, dblblindworkshop]{neurips_2026}
\makeatletter
\newcommand{\anon}[2][Anonymous]{%
  \if@neuripsfinal
    #2%
  \else
    \textcolor{red}{#1}%
  \fi
}
\makeatother

\usepackage[utf8]{inputenc} 
\usepackage[T1]{fontenc}    
\usepackage{hyperref}       

\usepackage{url}            
\usepackage{booktabs}       
\usepackage{longtable}      
\usepackage{amsfonts}       
\usepackage{nicefrac}       
\usepackage{microtype}      
\usepackage{xcolor}         

\usepackage[
style=numeric, 
urldate=long, 
backend=biber,
sortcites=true,
sorting=none,
]{biblatex}
\bibliography{own.bib}
\usepackage{xpatch}
\AtEveryBibitem{\ifentrytype{online}{%
}
{%
    \clearfield{url}%
    \clearfield{urlyear}%
    \clearfield{urlmonth}%
    \clearfield{urlday}%
  }%
}
\xpatchbibdriver{online} 
  {\printtext[parens]{\usebibmacro{date}}}
  {\iffieldundef{year}
    {}
    {\printtext[parens]{\usebibmacro{date}}}}
  {}
  {\typeout{There was an error patching biblatex-ieee (specifically, ieee.bbx's @online driver)}}
\DeclareFieldFormat{urldate}{Accessed: #1} 

\usepackage[inline]{enumitem} 

\usepackage{appendix}        
\usepackage{graphicx}        
\usepackage{amsmath}         
\usepackage{multirow}        
\usepackage{printlen}
\usepackage[nolist, nohyperlinks]{acronym} 

\newacro{LRP}{Layer-wise relevance propagation}
\newacro{ECG}{electrocardiogram}
\newacro{EEG}{electroencephalography}
\newacro{XAI}{explainable artificial intelligence}
\newacro{BIDS}{brain imaging data structure}
\newacro{BERT}{bidirectional encoder representations from Transformers}
\newacro{GPT}{generative pre-trained Transformer}
\newacro{FM}{foundation model}
\newacro{fMRI}{functional Magnetic Resonance Imaging}
\newacro{LaBraM}{large brain model}
\newacro{NN}{neural network}
\newacro{MHA}{multi-head attention}
\newacro{SDP}{scaled dot-product}
\newacro{RNN}{recurrent neural network}
\newacro{LSTM}{long short-term memory}
\newacro{NLP}{natural language processing}
\newacro{PPG}{photoplethysmography}
\newacro{iEEG}{intracranial Electroencephalogram}
\newacro{PET}{positron emission tomography}
\newacro{SHAP}{SHapley Additive exPlanations}
\newacro{MA}{masked autoencoding}
\newacro{DNN}{deep neural network}
\newacro{DTD}{deep Taylor decomposition}
\newacro{CFA}{cardiac field artifact}
\newacro{AUROC}{area under receiver operator characteristic curve}
\newacro{CSP}{common spatial patterns}
\newacro{LDA}{linear discriminant analysis}
\newacro{ReLU}{rectified linear unit}
\newacro{CSP-LDA}{common spatial patterns + linear discriminant analysis}
\newacro{EMG}{electromyogram}
\newacro{ALS}{amyotrophic lateral sclerosis}
\newacro{LLM}{large language model}
\newacro{VR}{virtual reality}
\newacro{JEPA}{joint-embedding predictive architecture}
\newacro{SSL}{self-supervised learning}
\newacro{CT}{computed tomography}
\newacro{MRI}{magnetic resonance imaging}
\newacro{CNN}{convolutional neural network}
\newacro{MLP}{multi-layer perceptron}
\newacro{BAC}{balanced accuracy}
\newacro{AUROC}{area under receiver operating characteristic curve}
\newacro{RQ}{research question}
\newacro{EOG}{electrooculogram}
\newacro{AOPC}{area over perturbation curve}
\newacro{MORF}{most relevant first}
\newacro{GXI}[G$\times$I]{Gradient $\times$ input}

\title{From Clever Hans to Hypotheses: Interpreting EEG Foundational Transformers}
\workshoptitle{Foundation Models for the Brain and Body}

\renewcommand{\thefootnote}{\fnsymbol{footnote}}
\author{
    Justus Meyer~zu~Bexten \footnotemark[1]\hspace{1.2mm}\footnotemark[2]\\
    \texttt{justus.bexten@uni-leipzig.de}
    \And
    Nico Scherf \footnotemark[1]\hspace{1.2mm}\footnotemark[2]\\
    \texttt{nscherf@cbs.mpg.de}
    \AND
    Bogdan Franczyk \footnotemark[1]\hspace{1.2mm}\footnotemark[3]\\
    \texttt{franczyk@wifa.uni-leipzig.de}
    \And
    Simon M. Hofmann \footnotemark[2]\hspace{1.2mm}\footnotemark[4]\\
    \texttt{simon.hofmann@cbs.mpg.de}
}

\begin{document}

\maketitle

\begin{abstract}
Emerging \acp{FM} in \ac{EEG} promise a path for deep learning in diagnostics and brain-computer interfaces despite data scarcity, yet their opaque nature remains a barrier to wider adoption. We investigate attention-aware \ac{LRP} as a post-hoc attribution method for \ac{EEG}-\acp{FM}, extending its use on \ac{CNN}-based \ac{EEG} models to \ac{EEG}-\acp{FM}. We find that \ac{LRP} can both verify \ac{EEG}-\ac{FM} decisions and surface novel, biologically plausible hypotheses from them. In motor imagery, it unmasks "Clever Hans" behavior where models prioritize task-correlated ocular signals over the intended motor correlates. In a naturalistic paradigm for affect prediction, it reveals a recurring reliance on a central electrode cluster, suggesting a candidate sensorimotor signature of arousal. 
Despite heatmap ambiguity, our results position \ac{LRP} as a tool for verifying and exploring EEG-FMs as these models mature.
\end{abstract}

\acresetall
\renewcommand{\thefootnote}{\arabic{footnote}}
\setcounter{footnote}{0}

\section{Introduction}
Large-scale pre-training promises an effective solution for deep modeling on \ac{EEG}-data \cite{liuEEGFoundationModels2026, jiangLargeBrainModel2023, wangCBraModCrissCrossBrain2025}, in spite of particularly pronounced data scarcity.
If \Ac{EEG} \acp{FM} deliver the promised generalization, the impact could be substantial (e.g., for speech prostheses \cite{cardAccurateRapidlyCalibrating2024}).
Early \ac{EEG} \acp{FM} have been trained on thousands of hours of recordings (e.g., >60,000~h \cite{ouahidiREVEFoundationModel2025}), and their underlying ‘understanding’ of EEG may prove increasingly informative. 
However, their decision processes are opaque.
To mitigate the black box problem of deep learning, feature attribution methods assign relevance scores to input features, identifying which signal components drive a model's decision.
\ac{LRP} \cite{bachPixelWiseExplanationsNonLinear2015, montavonExplainingNonlinearClassification2017, montavonLayerWiseRelevancePropagation2019} is a computationally efficient attribution method which offers faithful attributions across layers at high resolution.
\ac{LRP} is therefore well-suited to probe EEG FM decisions, both for verification and exploratory analysis.
Previous applications of  \ac{LRP} on \ac{EEG} were based on \acp{CNN} and \acp{MLP}, while \acp{FM} mostly rely on Transformers \cite{vaswaniAttentionAllYou2017}. Attention-aware \ac{LRP} \cite{achtibatAttnLRPAttentionAwareLayerWise2024}, allows us to bridge this gap.
We make three contributions:

\textbf{(1)} We are amongst the first to demonstrate that \textbf{\ac{LRP} produces plausible attributions on foundational EEG Transformers}, validated against a task with known ground truth (cardiac signal detection).\\
\textbf{(2)} Building on this, we \textbf{detect spurious behavior in \ac{EEG}-\acp{FM}}, revealing reliance on eye-movement related signals in motor imagery models.\\
\textbf{(3)} Finally, in the absence of intuition for \ac{EEG}, we characterize \textbf{unique promises and limits of \ac{LRP} on \ac{EEG}-\acp{FM}}: consistent and biologically plausible arousal patterns suggest \textbf{novel hypotheses}, pointing the way \textbf{towards learning from our learning machines.} However heatmap interpretation remains \textbf{inherently  ambiguous}.

\paragraph{Background}
In the domain of \ac{EEG}, \ac{FM} architectures typically combine temporal convolutional layers \cite{lecunGradientbasedLearningApplied1998} for patch extraction with Transformer-based attention layers \cite{vaswaniAttentionAllYou2017}.
Though first 'promptable' approaches exist \cite{jiangNeuroLMUniversalMultitask2024}, most current models, including the \ac{LaBraM} \cite{jiangLargeBrainModel2023}, REVE \cite{ouahidiREVEFoundationModel2025}, and CBraMod \cite{wangCBraModCrissCrossBrain2025}, utilize \ac{MA} objectives during pretraining.
Notable exceptions are the contrastive \ac{EEG} model BIOT \cite{yangBIOTBiosignalTransformer2023} and the recent EEG-DINO \cite{wangEEGDINOLearningEEG2026}. According to a benchmark by Liu et al. \cite{liuEEGFoundationModels2026}, these early \acp{FM} still rely heavily on full fine-tuning and struggle to consistently outperform specialized models trained from scratch.
In the more recent NeuralBench benchmark \cite{banvilleNeuralBenchUnifyingFramework2026a}, \ac{EEG}-\acp{FM} largely outrank specialized models. 
REVE \cite{ouahidiREVEFoundationModel2025}, the model with the largest pretraining corpus to date, scores first.
\ac{LaBraM} \cite{jiangLargeBrainModel2023} follows with the second best average rank.\\
\Ac{LRP} was first applied to \ac{EEG} by Sturm et al. \cite{sturmInterpretableDeepNeural2016} using a simple 2-layer network. Subsequent \ac{EEG} research has largely applied \ac{LRP} to single-task \acp{CNN} (e.g., \cite{zhouInterpretableSleepStage2024, nouriDetectionADHDDisorder2023}).
Beyond \ac{LRP}, other \ac{XAI} techniques have addressed epileptic seizure prediction \cite{khanExplainableFuzzyDeep2024} and schizophrenia diagnosis \cite{almadhorInterpretableXAIDeep2025}, while Nam et al. \cite{namFeatureSelectionBased2023} demonstrated that \ac{LRP} can actively improve model performance via guided feature selection.
Sutjatha Ravindran et al. have evaluated various propagation-based attribution methods in a simulation experiment \cite{sujatharavindranEmpiricalComparisonDeep2023}. While these studies were based on \acp{CNN} and \acp{MLP}, \ac{LRP} emerged as a suitable attribution method. Concurrent work \cite{sircaAccuracyRobustnessInterpretability2026} also applies AttnLRP to \ac{EEG}-\acp{FM}, proposing plausible attributions as a quality indicator beyond raw performance.
Our focus on Clever Hans behavior and hypothesis generation instead considers what explainable \acp{FM} can reveal about \ac{EEG}.


\section{Methodology and experiments}
\label{sec:methodology}

We apply the recent attention-aware \ac{LRP} \cite{achtibatAttnLRPAttentionAwareLayerWise2024} to the \ac{LaBraM} model \cite{jiangLargeBrainModel2023}, which shares architectural characteristics with many current \ac{EEG}-\acp{FM}. 
To address the intrinsic multidimensionality of both attributions and \ac{EEG} data, we devise a concise visualization for attribution heatmaps.
Through adequate controls, we ensure the isolation of \ac{FM}-specific behavior.

\paragraph{Isolating the influence of pretraining}
\label{paragraph:Isolating the influence of pretraining}
To isolate behavior specific to the pretrained model, we analyze three common downstream configurations:
The standard \textit{finetuned} approach, a \textit{frozen} variant where only the classification head is trained, and a \textit{from scratch} baseline using random initialization.
Performance is also measured. To account for variability due to random training effects such as dropout, initializations, and data sampling, each configuration is trained 5 times per task.


\paragraph{Extracting decision patterns by aggregation and visualization}
\label{paragraph:reporting_and_interpreting_attribution_patterns}
Subsequent to an inference decision, LRP generates attribution scores for every input time point in every channel. 
To extract understandable decision strategies from this large volume of data, we aggregate these multidimensional attributions across all samples and time steps to reveal spatial patterns, or across channels to isolate temporal patterns. \autoref{subsec:attribution_aggregation_process} details this process step by step.

\paragraph{The explainable EEG-FM: LaBraM with LRP}
We use the public 'base' version of the \ac{EEG}-\ac{FM} \ac{LaBraM} \cite{jiangLargeBrainModel2023}. Its 5.8~mio. parameters are distributed across patch-wise convolutional layers, and encoder-attention blocks, a constellation that has prevailed in more recent models like CBraMod \cite{wangCBraModCrissCrossBrain2025} or EEG-DINO \cite{wangEEGDINOLearningEEG2026}. "Attention-aware LRP" \cite{achtibatAttnLRPAttentionAwareLayerWise2024} allows us to apply \ac{LRP} \cite{montavonExplainingNonlinearClassification2017,montavonLayerWiseRelevancePropagation2019, bachPixelWiseExplanationsNonLinear2015}
to \ac{LaBraM}. \Ac{LRP} propagates 'relevance' back through the network layers from an output logit, using gradient rules in linear computations and altered gradient rules for non-linear computations \cite{arrasCloseLookDecompositionbased2025}. It is able to provide attributions to inputs on a fine-grained level as well as to intermediate layers if required (in contrast to Attention Rollout \cite{abnarQuantifyingAttentionFlow2020} and \ac{SHAP} \cite{lundbergUnifiedApproachInterpreting2017}). 
Details on the \ac{LRP}-configuration and \ac{LaBraM}'s architecture can be found in \autoref{sec:details on training and evaluation}.

\paragraph{Experiments: Questions and approach}
\label{sec:experiments}
We explore three guiding questions in the context of \ac{EEG}-Transformers: (i) whether \ac{LRP} produces plausible attributions on this class of model; (ii) whether it surfaces the kinds of insights typical of \ac{XAI} use cases, such as exposing spurious decision strategies; (iii) and whether combining \ac{XAI} with \ac{EEG}-\acp{FM} is a productive direction for exploratory research. To address these directions, we apply our methodology to both established and experimental EEG paradigms of increasing complexity.
A validation experiment based on the \ac{CFA} and experiments in the established paradigm of motor imagery/execution speak to the first two; experiments on the more challenging prediction of affective states speak to the third.



\subsection{Data and tasks}
\label{subsec:Datasets}
The following datasets are used for our experiments, train/test/validation splitting was conducted on the subject level. Preprocessing was done in closest possible adherence to \ac{LaBraM}s protocol. \autoref{sec:data acquisition and cohort details} contains further details on this, as well as on cohort, acquisition, and synthesis of labels.

\paragraph{R-Peak detection --- validation target} As a verification experiment for our \ac{LRP} configuration, we predict R-Peak occurrence from raw \ac{EEG}: Particularly in electrodes closer to the heart, each heartbeat induces a \ac{CFA} in the signal \cite{dirlichCardiacFieldEffects1997}, yielding a task with a known expected decision strategy. The experiment is restricted to the \emph{finetuned} configuration.

\paragraph{PhysioMI L/R --- established paradigm}
\label{par:PhysioMI}
To conduct experiments in a well understood paradigm, we derive a subset of the PhysioNet \cite{goldbergerPhysioBankPhysioToolkitPhysioNet2000} EEG Motor Movement/Imagery Dataset\footnote[1]{Available under the ODC Attribution License at \url{https://physionet.org/content/eegmmidb/1.0.0/}} \cite{schalkEEGMotorMovement2009}. 
It contains 64-electrode \ac{EEG} recordings during real or imagined directional movements elicited by the display of directional targets on a screen.
We include only runs for task 1 (left/right fist closing) and task 2 (imagined left/right fist closing), referring to this subset as PhysioMI L/R. 
We predict directional intent (left or right) of imagined or executed motion.

\paragraph{AffectiveVR --- challenging setting }
\label{par:AffectiveVR}
The \emph{AffectiveVR} dataset contains 64-electrode EEG-recordings, physiological recordings, and subjective feedback on the core affective states \cite{russellCircumplexModelAffect1980} arousal ('elicited exitedness') and valence ('elicited positivity/negativity') during a virtual reality experience. We predict these affective states separately, in a binary manner (High / Low).
The dataset is not yet public but we refer to related work by the authors \anon[(anonymized citations)]{\cite{fourcadeLinkingBrainHeart2024, fourcadeAffectTrackerRealtimeContinuous2025}}.

\subsection{Training and evaluation}
\label{subsec:training and evaluation}
We train our models using the AdamW optimizer \cite{loshchilovDecoupledWeightDecay2018}, using a cosine annealing learning rate schedule \cite{loshchilovSGDRStochasticGradient2017}, L2-regularization, label smoothing, early stopping based on validation loss, dropout and stochastic depth \cite{huangDeepNetworksStochastic2016} in close adherence with the setup used in the original \ac{LaBraM} publication \cite{jiangLargeBrainModel2023}.
For \textit{frozen} configurations, we use an MLP classification head, and linear heads for all other configurations.
We compare performances to \ac{CSP-LDA}, an established method for binary discrimination in \ac{EEG}-analysis \cite{ramoserOptimalSpatialFiltering2000, blankertzOptimizingSpatialFilters2008}.
For details on hyperparameters, compute, training, and evaluation, refer to \autoref{sec:hyperparam_search} and \autoref{sec:details on training and evaluation}, respectively.

\section{Results and discussion}
\label{sec:results}
\subsection{Classification performance}
In \textbf{R-Peak detection}, a \ac{BAC} of approximately 75\% is reached. This is well above the 50\% baseline expected of a non-discriminative classifier and \emph{sufficient for our verification purpose}.
For both \textbf{AffectiveVR} tasks, \ac{BAC} measures around 55\% are \emph{markedly better than a random baseline} (which at a median split is expected around 50\%), and \emph{on par with \ac{CSP-LDA}}, though AUROC scores tend to be slightly higher than for \ac{CSP-LDA}.
Difference in performance between finetuned and from scratch configurations is negligible.
While the low binary accuracy may be construed as a lack of training success, previous subjective arousal state prediction \cite{hofmannDecodingSubjectiveEmotional2021}, yielded a performance of approx. 60\% \ac{BAC}.
Given the different constraints of this aforementioned work (intra-subject training, elimination of ambiguous samples), we consider the performance within the expected range for this experimental prediction task.
The attribution patterns we observe for this task are thus connected to a successful prediction strategy for its subjective and partially ambiguous prediction target.
\textbf{PhysioMI L/R }generalization performance is mostly \emph{in the range of random guessing}, except for the finetuned configuration:
Two out of five \emph{finetuning runs displayed strong generalizable performance} of over 70\% \ac{BAC}, while the other three generalized similarly well as the other configurations.
Training loss did converge for all runs, pointing towards overfitting in most runs.
>70\% \ac{BAC} is approximately consistent with performances displayed on motor imagery tasks by finetuned \acp{FM} in a prior benchmark \cite{liuEEGFoundationModels2026}.
Our analysis of attribution patterns reveals aspects of the PhysioMI experiment design which may be connected to the dominant non-generalizable decision strategy. For further consideration, \autoref{tab:performance_metrics} lists all performance metrics by configuration and task.

\subsection{Attribution patterns - from validation to exploration}
\label{subsec: attributionpatterns results}
\begin{figure}
    \centering
    \includegraphics[]{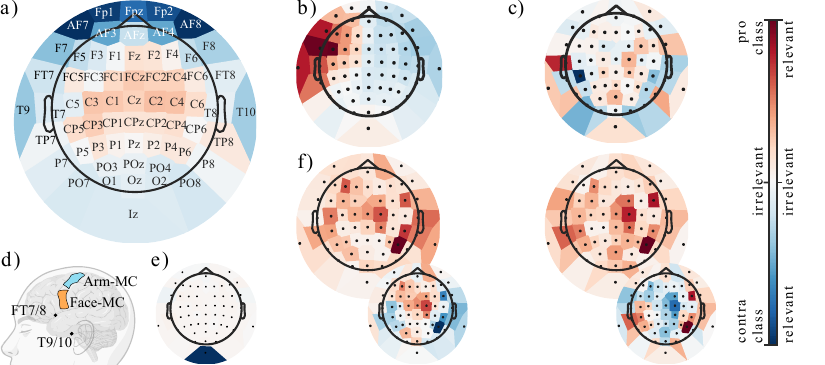}
    \caption{Exemplary attribution patterns from various tasks, Voronoi-cell colored to avoid confusion with activity-related maps. Right of the colorbar is how we interpret relevance heatmaps given sign flips. \textbf{a)} Attribution pattern similarly found in many non-generalizing runs for PhysioMI L/R, likely overfitted on eye-movement artifacts. 
    Generalization performance is close to random guessing.
    \textbf{b)} Second of two distinct attribution patterns for PhysioMI L/R. Test \ac{BAC} of this Run: 74.99\%. 
    \textbf{c)} Typical attribution pattern for valence state prediction. 
    \textbf{d)}\protect\footnotemark[2] Temporal electrode positions in relation to approximate arm/hand- and face/eye-related motor cortex areas according to \cite{gordonSomatocognitiveActionNetwork2023}.
    \textbf{e)} Typical R-Peak attribution pattern, all relevance concentrated at the Iz electrode.
    \textbf{f)} Absolute patterns from different arousal prediction runs, with oriented originals to their bottom right. The patterns are almost identical apart from sign flip. }
    \label{fig:attribution_patterns_dense}
\end{figure}
\footnotetext[2]{Created in BioRender. \anon{Meyer zu Bexten, J. (2026) https://BioRender.com/0qa0ei2}}

During our analysis of attribution heatmaps, we qualitatively identified task-specific patterns which re-occur across various runs and model configurations (statistical assessment in \autoref{subsec:quantitative-electrode-consistency}), implying some dominant learned model behaviors. We analyze attribution maps and consider implied decision strategies.
Finally, we interpret their biological plausibility in light of previous literature, and address possible novelties and ambiguities.
While patterns occur consistently, they are sometimes \textbf{sign-inverted} resulting in reversed coloring in their visualizations (see \autoref{fig:attribution_patterns_dense} f).
Despite the observed sign inversions, the absolute magnitude of relevance reveals stable, task-specific patterns, indicating that the spatial structure of the decision process remains consistent regardless of the attribution's sign. 

\paragraph{Unambiguous validation: R-Peak detection} The attribution pattern for R-Peak detection is consistent across runs apart from sign inversion,
 with relevance concentrated at the lower-lateral Iz electrode (see \autoref{fig:attribution_patterns_dense} e). \Acp{CFA} exhibit the highest amplitude in electrodes closest to the heart \cite{dirlichCardiacFieldEffects1997}, and Iz is the closest in AffectiveVR's montage. This fully aligns with the expected decision strategy, lending confidence to attention-aware \ac{LRP} in the analyses to follow.

\paragraph{Discovering Clever-Hans behavior: PhysioMI L/R}
\label{par:physiomi_attr_analy}
We observe a sharp distinction in XAI patterns: successful models produce lateralized, temporo-frontal attributions (see \autoref{fig:attribution_patterns_dense} b), while non-successful models focus heavily on ocular regions (see \autoref{fig:attribution_patterns_dense} a). 
The poor generalization of non-successful models may stem from their reliance on ocular artifacts, possibly caused by the lateralized visual cues (e.g., a left-side marker on the screen for left-fist clenching) of the PhysioMI experiment,  as a non-generalizing shortcut. 
While the lateralized nature of successful models aligns with our expectations for a left/right motor task, we notice that the outermost temporal electrodes which our model focuses are closer to regions that Gordon et al. \cite{gordonSomatocognitiveActionNetwork2023} associate with eye movements than they are to hand-related regions.
Thus, even successful models may rely on ocular movement related neural code rather than the intended neural signatures of hand-motor imagery (ablation in \autoref{subsec:physiomi-ablation}).
Using \ac{LRP}, we are able to uncover this 'Clever Hans' behavior.

\paragraph{Familiar and novel patterns: arousal state prediction}
On AffectiveVR, we observe a recurring attribution pattern across configurations: strong relevance at the P6 electrode (right parietal-occipital region), consistent with its established association with subjective arousal \cite{hofmannDecodingSubjectiveEmotional2021, lettieriEmotionotopyHumanRight2019}. We also consistently find opposing relevance in a central electrode cluster (C2, Cz, FC2, CP2; see \autoref{fig:attribution_patterns_dense}f), overlying the sensorimotor cortex, a region associated with the \emph{readiness potential} preceding voluntary hand movements\cite{kornhuberHirnpotentialaenderungenBeiWillkuerbewegungen1965, gerbrandtDistributionHumanAverage1973}. It is known to be modulated by emotional context \cite{oliveiraPreparingGraspEmotionally2012, senkowskiEmotionalFacialExpressions2011}, suggesting two interpretations: the model detects the neural signature of thumb movements used for continuous feedback required in the task, or an arousal-specific state of motor preparedness. This central pattern is absent in valence prediction (see \autoref{fig:attribution_patterns_dense}c), which uses identical thumb-based feedback, thus pointing toward an arousal-specific sensorimotor signature.
Sensorimotor electrodes have appeared in prior arousal research \cite[Discussion]{hofmannDecodingSubjectiveEmotional2021}, \cite[Table 4]{koelstraDEAPDatabaseEmotion2012}, \cite[Figure 4c]{fourcadeLinkingBrainHeart2024}, but their link has not been studied explicitly.
In future studies, \ac{XAI} applied to more capable EEG FMs may surface further candidate hypotheses of this kind.

Valence state prediction yielded distinctly consistent attribution patterns, without emphasizing the center electrode cluster. This strengthened our aforementioned candidate hypothesis.
However, only one electrode relevant to valence appeared in previous literature, and most were close to neck- and face musculature (see \autoref{subsec:extended_valence_attribution_analyisis} for specifics).
This prevented a conclusive interpretation, and highlights the particular impact of heatmap ambiguity on EEG data.
In contrast to XAI applications for natural images or language, disambiguation by intuition proves difficult in the EEG domain.


\section{Conclusion}
\label{sec:conclusion}
\paragraph{\ac{EEG}-\acp{FM} require \ac{XAI}: \ac{LRP} can provide it} In our R-Peak and PhysioMI L/R experiments, we show that attention-aware \ac{LRP} transfers successfully to \ac{EEG}-\ac{FM} architectures. It yields attributions consistent with known neurophysiology and, moreover, \emph{explains overfitting} and \emph{reveals shortcut learning} (a problem that has undermined prior work, e.g., \cite{liPerilsPitfallsBlock2021}). 
The latter two are particularly valuable in verifying \ac{EEG} experiments, or even benchmark integrity.

\paragraph{Towards exploration through \acp{FM}}Our work additionally demonstrates that \ac{XAI} applied to \ac{EEG} \acp{FM} can surface novel, plausible hypotheses on how brain functions relate to predicted variables.
Though bounded by ambiguity and \ac{FM} quality, this instance of hypothesis discovery points towards a productive direction for \ac{EEG} \ac{FM}-based research beyond prediction performance.

\paragraph{Limitations}
\label{par: limitations}
Three caveats bound our findings.
\emph{First}, \ac{LRP} attributions in low-intuition domains like EEG remain difficult to interpret, and the visual nature of heatmap inspection (including in this work) is susceptible to subjective bias. Verifying a model's decision strategies will remain a research-intensive task.
\emph{Second}, our use of \ac{LRP} rests on prior comparisons of attribution methods conducted on non-Transformer EEG models \cite{sujatharavindranEmpiricalComparisonDeep2023}, and validation experiments (\autoref{subsec: attributionpatterns results} and \autoref{subsec:faithfulness_eval}). Broader \ac{XAI} comparisons on EEG FMs are a natural next step.
\emph{Third}, the absence of stable performance gains on AffectiveVR and PhysioMI L/R, as well as recent benchmarks \cite{liuEEGFoundationModels2026} suggests that the FM's "understanding" of EEG is not yet leveraged consistently. Until this changes, the discovery potential of our approach is constrained by the underlying \ac{EEG} \acp{FM}.

\paragraph{Outlook} 
\label{subsec:outlook}
Open data efforts and research competitions \cite{aristimunhaEEGFoundationChallenge2025} substantiate an optimistic outlook for \ac{EEG} \acp{FM}, and \ac{LRP} can contribute to this trajectory by informing model development and evaluation. As downstream applications become feasible, verification of model decisions will become increasingly relevant, and \ac{LRP} is well positioned to assist in this. Two extensions are particularly promising. Transforming attribution scores into the frequency domain would render raw  \ac{EEG}-\ac{FM} decisions interpretable for domain experts accustomed to spectral \ac{EEG} analysis. Interactive visualization tools would complement this with finer-grained inspection of individual decisions.
Furthermore, by highlighting the most informative signal components, attribution scores can facilitate a more targeted expert analysis of large-scale EEG data.
Our work provides a methodological foundation and reference for \ac{XAI} development along these directions.

\section*{Acknowledgments}
\anon[Anonymized]{J.~M.~z.~B., S.~M.~H. and N.~S. are supported by BMFTR (Federal Ministry of Research, Technology and Space) through ACONITE (01IS22065). 
J.~M.~z.~B., N.~S. and B.~F. are further supported by the Federal Ministry of Research, Technology and Space of Germany and by Sächsische Staatsministerium für Wissenschaft, Kultur und Tourismus in the program Center of Excellence for AI-research „Center for Scalable Data Analytics and Artificial Intelligence Dresden/Leipzig“, project identification number: ScaDS.AI.
J.~M.~z.~B. was further supported by the BMFTR through a scholarship of DAAD project 57616814 (SECAI, School of Embedded and Composite AI) as part of the program Konrad Zuse Schools of Excellence in Artificial Intelligence.
Computations were performed on the HPC systems \anon[anonymized nvidia cluster]{Raven} and Viper at the Max Planck Computing and Data Facility}

\renewcommand*{\bibfont}{\small}
\printbibliography

\begin{appendices}
\setcounter{table}{0}
\setcounter{figure}{0}
\renewcommand{\thetable}{A\arabic{table}}
\renewcommand{\thefigure}{A\arabic{figure}}

\section{Further details on data}
\label{sec:data acquisition and cohort details}
\paragraph{R-Peak detection}
Using R-Peak annotations from the AffectiveVR dataset (introduced below), we construct a binary target signal at \ac{EEG} resolution by setting a 20~ms window around each peak to 1 and the rest to zero. This results in $t_{in}=t_{out}=800$ logits. To limit storage requirements, we thus subsample 2 positive and 2 negative logits for attribution (see \autoref{paragraph:reporting_and_interpreting_attribution_patterns}). 

\paragraph{PhysioMI L/R}
We exclude three out of 109 participants for which sampling frequency and event timings deviated from given specifications. 
The subjects are divided into training, validation, and test sets randomly with an 80:10:10 ratio.
We extract 4-second epochs/windows corresponding to the motor tasks.
The objective is to classify the directional intent (left or right) across both execution and imagery trials, which amounts to $t_{out}=1$. Event annotations are used as labels.

\paragraph{AffectiveVR}
The recording setup for EEG and physiological data in a naturalistic setting shares similarities to \anon[anonymized study]{\cite{fourcadeLinkingBrainHeart2024}}.
Stimuli and emotional feedback mechanism are identical to those described in \anon[anonymized study]{\cite[Study 2]{fourcadeAffectTrackerRealtimeContinuous2025}}. 
We will refer to this dataset as \emph{AffectiveVR}.
Participants were placed in a swivel chair and shown \ac{VR} scenes ranging from ghastly and dimly lit to serene and including fluffy creatures.
Using a touchpad controller, they subjectively reported their imminent emotional state according to James A. Russel's model of affect \cite{russellCircumplexModelAffect1980}, along the axis of arousal and valence (which could colloquially be described as elicited exitedness and positivity/negativity). Additionally, a 3-electrode \ac{ECG} and a 64-electrode \ac{EEG} were recorded along with further physiological signals and information (including eye and head tracking, galvanic skin response, respiratory measurements, \ac{PPG}, and a questionnaire). R-Peak timings were detected automatically and manually corrected using NeuroKit2's default method \cite{brammerBiopeaksGraphicalUser2020}.
The study included 47 participants, consisting of 24 women, 22 men and one non-binary person.
Mean age was 27.10 years, with a standard deviation (SD) of 5.37. 4 participants were excluded by the dataset providers for quality reasons, 2
further participants are excluded in our analysis due to unusual affective feedback,
namely the reporting of the same state throughout most of the experiment. This would have made a meaningful median split impossible.
AffectiveVR is provided to us with the following preprocessing steps already applied:
Data was cleaned by applying a 50~Hz power line filter and 4th-order Butterworth bandpass filters (0.5--30~Hz for ECG. 0.1--45~Hz for EEG) using the Python toolbox NeuroKit2(v0.2.7. \cite{makowskiNeuroKit2PythonToolbox2021}).
Arousal and valence feedback were linearly interpolated where values were missing (i.e., when the participants took their thumb off the touchpad controller), and then downsampled to 1~Hz by taking the mean value over 1-second windows. 
From the 41 quality-controlled participants, we sample validation ($n_{men}=n_{women}=2$ and hold-out test ($n_{men}=n_{women}=4$) subsets by gender, and use the remaining subjects for training ($n_{men}=10, n_{women}=18, n_{non-binary}=1$),  prioritizing a balanced evaluation. 
Every logit produced by a model can produce an attribution heatmap through \ac{LRP}. To not further increase the axes of our analysis, we operationalize affective states into binary classes, yielding only a single logit per classification decision. They are present as a resampled 1 Hz signal, and due to it's subjective nature, should be regarded as relative for each participant.
We therefore choose the intra-subject median as a binarization boundary. We label a 1~Hz time-step 'high' if it exceeds this median or 'low' otherwise. Training examples are created through a rolling window approach with 4-s windows and 1-s stride.

\paragraph{EEG preprocessing}
\label{par:preprocessing}
To better adhere to \ac{LaBraM} specifications, we apply the following preprocessing:
To PhysioMI L/R, we apply a 0.1 to 75 Hz FIR-bandpass filter, and a 50 Hz notch filter to eliminate power-line interference. This step is omitted for AffectiveVR, as it is already filtered. We re-reference EEG to the average across electrodes, and resample to 200~Hz.
These additional preprocessing steps are performed using MNE Python toolkit (v1.4.2; \cite{larsonMNEPython2023}).
For parallelization and loading, we leverage HuggingFace Datasets (v3.6.0. \cite{Datasets}).
At this resolution, the 4-s input windows in our tasks amount to $t_{in}=800$ input time steps.

\section{Details on training and evaluation}
\label{sec:details on training and evaluation}

\begin{table}[tbhp]
\centering
\caption{Hyperparameters used in training.}
\label{tab:hyperparam_search_results}
\begin{tabular}{@{}lllllllllll@{}}
\toprule
Task                                    & Conf.       & $B$   & Dropout  & L2-Reg.   & $w_{+}$ & $n_{h}$ & $n_{l}$ & $lr$       & $e_{wu}$ & $lr_{wu}$ \\ \midrule
\multirow{2}{*}{R-Peak}       & f.s. & 32  & 31.3\%  & 7.6E-05 & 6.47    & -         & -         & 5.0E-04 & 0         & -          \\
                              & ftn.     & 32  & 26.3\%  & 1.5E-03 & 6.33    & -         & -         & 5.0E-04 & 0         & -          \\ \midrule
\multirow{3}{*}{Arousal}      & f.s. & 128 & 89.5\%  & 3.4E-06 & -       & -         & -         & 5.0E-04 & 0         & -          \\
                              & ftn.     & 128 & 90.7\%  & 4.7E-04 & -       & -         & -         & 5.0E-04 & 0         & -          \\
                              & frz.     & 256 & 42.3\%  & 1.5E-03 & -       & 64        & 2         & 5.0E-04 & 0         & -          \\ \midrule
\multirow{3}{*}{Valence}      & f.s. & 64  & 82.1\%  & 3.6E-06 & -       & -         & -         & 5.0E-04 & 0         & -          \\
                              & ftn.     & 128 & 77.0\%  & 2.3E-04 & -       & -         & -         & 5.0E-04 & 0         & -          \\
                              & frz.     & 256 & 37.0\%  & 2.1E-03 & -       & 64        & 2         & 5.0E-04 & 0         & -          \\ \midrule
\multirow{3}{1.2cm}{PhysioMI L/R} & f.s. & 32  & 11.4\%  & 1.1E-03 & -       & -         & -         & 3.4E-05 & 0         & -          \\
                              & ftn.     & 32  & 69.2\%  & 8.5E-04 & -       & -         & -         & 7.6E-05 & 8         & 5.9E-06    \\
                              & frz.     & 128 & 10.1\%  & 6.9E-06 & -       & 64        & 1         & 1.2E-05 & 0         & -          \\ \bottomrule
\end{tabular}
\end{table}

We train  all \ac{LaBraM}-configurations using the AdamW optimizer \cite{loshchilovDecoupledWeightDecay2018}.
We use a cosine annealing learning rate schedule \cite{loshchilovSGDRStochasticGradient2017} starting at learning rate $lr = \text{5E-04}$ without warmup. 
Through this approach, fine local minima become increasingly dominant in the course of the optimization, which amounts to an initially coarse and later fine adaption of parameters to the training objective.
In PhysioMI experiments, we opt to use linear warmup during finetuning, and determine warmup epochs $e_{wu}$ as well as warmup and start learning rate ($lr_{wu}$, $lr$) through hyperparameter tuning.

Additionally, to prevent overfitting on the training dataset, we use early stopping with a grace period of 10\% of the maximum epochs.
If the validation loss does not improve for this duration, training is stopped and the model is reset to the epoch with the highest validation balanced accuracy. 
Maximum epochs are set to 100 for AffectiveVR, and to 400 for PhysioMI to compensate for the smaller overall sample amount. 
We apply label smoothing with factor 0.1 \cite{szegedyRethinkingInceptionArchitecture2015}.
We tune hyperparameters for L2-regularization, batch size $B$, and a dropout hyperparameter that is used for both hidden-unit- and attention dropout, as well as the stochastic depth parameter. Stochastic depth dropout stochastically bypasses a layer through it's skip connection with likelihood linearly increasing from early layers to later ones, up to the aforementioned stochastic depth parameter (refer to \ac{LaBraM} code). While the authors don't comment on this, the implementation resembles the work of Huang et al. \cite{huangDeepNetworksStochastic2016}).

For frozen configurations, we also tune the number of hidden layers $n_\text{layers}$ and the number of hidden units per layer $n_\text{hidden}$ in the MLP classification head. In case of the R-peak prediction task, we also tune a weight $w_\text{peak}$ for class-weighted loss due to the great class imbalance. The hyperparameters used are displayed \autoref{tab:hyperparam_search_results}. The hyperparameter search outlined in (\autoref{sec:hyperparam_search}). 
As mentioned before, we include linear warmup in the PhysioMI finetuning experiment, and tune $lr$ for all PhysioMI experiments.
Random effects during training can affect performance and decision patterns. To estimate the impact of these, we repeat training and evaluation five times for each target variable.

For \ac{CSP-LDA}, we optimize the number of components by fitting on the training set and evaluating on the validation set using a grid search, which yields 12 components for affect prediction and 4 for PhysionMI L/R.
We use the \ac{CSP} implementation from MNE (v1.4.2.\cite{larsonMNEPython2023}) and the \ac{LDA} implementation from scikit-learn (v1.4.1.post1.\cite{pedregosaScikitlearnMachineLearning2011}). The variance for CSP-LDA is estimated via 1000-fold bootstrap rather than over five runs (denoted with $*$ in \autoref{tab:performance_metrics}).

\label{par:computational setup}
For each configuration and target, 5 training, evaluation and attribution runs were executed. 
This was done on the \anon[anonymized nvidia cluster]{Raven HPC cluster}
 of \anon[anonymized compute provider]{the 
Max Planck Society's Computing and Data Facility (MPCDF; Garching, Germany)}
on 1 NVIDIA A100 with 40~GB high bandwidth memory running PyTorch with Nvidia CUDA (v12.6). 
More details on the clusters can be found in the respective documentation \anon[anonmyized citation]{\cite{RavenUserGuide}}.  
 A single run terminated within approx. 1.5~h, with some runs terminating significantly earlier due to early stopping.
Including attribution, the five runs of one experiment typically terminated within 5~h.
All CSP-LDA experiments were executed on a consumer laptop with 8 CPU-cores and 24 GB of RAM, with a runtime of less than an hour, including the gridsearch for the optimal component count.

\paragraph{LaBraM Architecture}
\label{par:LaBraM_architecture_details}
We use the public 'base' version of \ac{LaBraM} \cite{jiangLargeBrainModel2023}, which has 5.8~mio. parameters. It was pretrained using a discretized \ac{MA} objective, on 2500~h of EEG recordings, and takes input as channel-wise input patches of 1~s.
Each patch is fed through 3 temporal convolutional blocks.
A 'CLS'-token is appended, and learned temporal- and channel-encodings are added to the patch representations. 
This is followed by 12 encoder attention layers with 10 attention heads each.
The original publication \cite{jiangLargeBrainModel2023} contains further details.

\paragraph{LRP-Configuration details}
\label{par:appendix_lrp_config}
For each logit produced for each sample in the test set, we create attribution scores using \ac{LRP}. We use the \ac{LRP} rules provided in the LXT Python library (v2.0, \cite{achtibatAttnLRPAttentionAwareLayerWise2024}) for Transformer components of \ac{LaBraM}, which are equivalent to epsilon-LRP.
By using the "efficient" variant of LXT, which utilizes the "modified Grad$\times$Input" perspective on LRP, the epsilon parameter is eliminated \cite{arrasCloseLookDecompositionbased2025}. 
For the early temporal convolutional layers in \ac{LaBraM}, we use the gamma rule with $\gamma=0.25$. For the input layer, we use the $w^2$ rule, which is suitable for unbounded input. For both, we use $\epsilon=1e-6$. 
We employ the zennit (v0.5.1, \cite{andersSoftwareDatasetwideXAI2023}) library  on these convolutional layers, since LXT focuses on Transformer-specific layers.


\section{Hyperparameter search}
\label{sec:hyperparam_search}

\begin{table}[h]
\centering
\caption{Hyperparameter search spaces. "Affective" refers to the valence/arousal tasks. }
\begin{tabular}{@{}llll@{}}
\toprule
Hyperparameter      & Search Space /   Values & Distribution   & Experiment(s)      \\ \midrule
\multicolumn{4}{c}{General Parameters}                                              \\
Batch Size          & \{32, 64, 128, 256\}    & Discr. Uniform & All                \\
Dropout Rate        & {[}0.1, 0.99{]}         & Uniform        & All                \\
Weight Decay        & {[}1e-6, 0.01{]}        & Log-uniform    & All                \\ \midrule
\multicolumn{4}{c}{Head Architecture (Fine-tuning)}                                 \\
Hidden Layers       & \{0, 1, 2, 3, 4\}       & Discr. Uniform & Frozen (All)       \\
Hidden Dimension    & \{8, 16, 32, 64\}       & Discr. Uniform & Frozen (All)       \\ \midrule
\multicolumn{4}{c}{Further   Experiment-Specific Parameters}                              \\
Binary Class Weight & {[}5, 8{]}              & Uniform        & R-Peak Detection   \\
Learning Rate       & {[}5e-6, 1e-4{]}        & Log-uniform    & PhysioMI L/R (All) \\
Warmup Epochs       & \{0, 1, …, 40\}         & Uniform        & PhysioMI L/R ftn.  \\
Warmup Start LR     & {[}1e-6, 1e-5{]}        & Log-uniform    & PhysioMI L/R ftn.  \\ \bottomrule
\end{tabular}
\label{tab:hyperparam_search_spaces}
\end{table}

For the hyperparameter search, we use the Optuna Python library (v4.3.0, \cite{akiba2019optuna}) with a Tree-Structured Parzen Estimator \cite{watanabeTreeStructuredParzenEstimator2023} for efficient hyperparameter sampling. 
After 10 warm up trials, we use median pruning to discard unpromising trials early, starting after the 10th epoch. 
This means that any trial that is beyond epoch 10 will be terminated if its objective metric is below the median of previous trials. Balanced accuracy is used as the objective metric for hyperparameter optimization, and is evaluated on the validation set.
For AffectiveVR experiments, We opt not to include the learning rate as a tuning hyperparameter, since pruning, while decreasing computational effort significantly, tends to eliminate 'late bloomers', i.e. trials that learn slowly but reach better performance later on.
For PhysioMI experiments, we opted to include learning rate in hyperparameter tuning, in spite of the aforementioned concerns.
Additionally, our utilization of a cosine annealing learning rate schedule already leads to the use of various learning rates during training.
We run 100 trials for each experiment.
Refer to \autoref{tab:hyperparam_search_spaces} for the hyperparameter search spaces.
Hyperparameter tuning for R-peak detection and arousal/valence prediction is executed on the \anon[anonymized AMD cluster]{Viper HPC cluster}
 of \anon[anonymized compute provider]{the Max Planck Society's Computing and Data Facility (MPCDF; Garching, Germany)}, 
on 1 AMD Instinct MI300A with 128GB high bandwidth memory.
During the experiment on valence prediction, memory restrictions were introduced, limiting memory during hyperparameter tuning on the cluster to 64GB high bandwidth memory on 1 AMD Instinct MI300A.
On the \anon[anonymized AMD cluster]{"Viper" cluster}, we use PyTorch with AMD ROCm (v6.3) for APU compatibility.
Due to a forced ROCM-update on 
\anon[anonymized AMD cluster]{Viper}, hyperparameter tuning for PhysioMI L/R experiments was conducted under the same conditions as training itself (see \autoref{par:computational setup}).
More information on the infrastructure can be found in the official documentation \anon[anonymized citation]{\cite{ViperGPUUserGuide, RavenUserGuide}}.

For CSP-LDA, we use the aforementioned grid search approach to tune number of components (see \autoref{sec:details on training and evaluation}).

\section{Supplements on model performance}
\label{sec:extended_reporting_perfomance}
Result reporting in the body of the paper focuses on particularly noteworthy results. \autoref{tab:performance_metrics} lists extensive performance metrics for all experiments from the main part. We also provide curves of validation-split metrics in the course of training in \autoref{fig:val_baccs}. Furthermore, we offer a brief cross-validation experiment on arousal prediction in \autoref{subsec:cross-validation}.

\begin{table}[hbtp]

\centering
\caption{Test performance (mean \(\pm\) SD) for each configuration and task. (*) see \autoref{sec:details on training and evaluation}}
\begin{tabular}{@{}lllll@{}}
\toprule
\multicolumn{1}{l}{Task}              & Model          & AUROC                  & B. Acc.                 & F1                     \\ \midrule
\multirow{1}{*}{R-Peak   detection}   & LaBraM finetuned & \textbf{0.830 ± 0.003} & 75.0\% ± 0.3            & \textbf{0.171 ± 0.004} \\ \midrule
\multirow{4}{*}{PhysioMI   L/R}       & LaBraM finetuned & \textbf{0.672 ± 0.143} & \textbf{62.5\% ± 10.7}  & \textbf{0.612 ± 0.138} \\
                                      & LaBraM from scratch  & 0.513 ± 0.026          & 51.0\% ± 1.8            & 0.434 ± 0.103          \\
                                      & LaBraM frozen   & 0.502 ± 0.005          & 50.2\% ± 1.0            & 0.204 ± 0.203          \\
                                      & CSP-LDA *      & 0.578 ± 0.020          & 57.6\% ± 2.0            & 0.563 ± 0.024          \\ \midrule
\multirow{4}{*}{Arousal Prediction}   & LaBraM finetuned & 0.562 ± 0.015          & 54.6\% ± 1.2\%          & 0.558 ± 0.039          \\
                                      & LaBraM from scratch  & \textbf{0.578 ± 0.006} & \textbf{55.6\% ± 0.7\%} & \textbf{0.585 ± 0.037} \\
                                      & LaBraM frozen   & 0.563 ± 0.004          & 54.5\% ± 0.3\%          & 0.566 ± 0.020          \\
                                      & CSP-LDA *      & 0.555 ± 0.004          & 55.5\% ± 0.5\%          & 0.550 ± 0.006          \\ \midrule
\multirow{4}{*}{Valence   Prediction} & LaBraM finetuned & \textbf{0.574 ± 0.012} & 55.2\% ± 0.7\%          & \textbf{0.564 ± 0.033} \\
                                      & LaBraM from scratch  & 0.562 ± 0.004          & 54.7\% ± 0.6\%          & 0.548 ± 0.019          \\
                                      & LaBraM frozen   & 0.532 ± 0.010          & 52.0\% ± 0.6\%          & 0.490 ± 0.083          \\
                                      & CSP-LDA *      & 0.557 ± 0.004          & \textbf{55.7\% ± 0.5\%} & 0.556 ± 0.006          \\ \bottomrule
\end{tabular}
\label{tab:performance_metrics}
\end{table}

\subsection{Arousal prediction with cross-validation}
\label{subsec:cross-validation}
\begin{figure}[thbp]
\centering
\includegraphics[width=\textwidth]{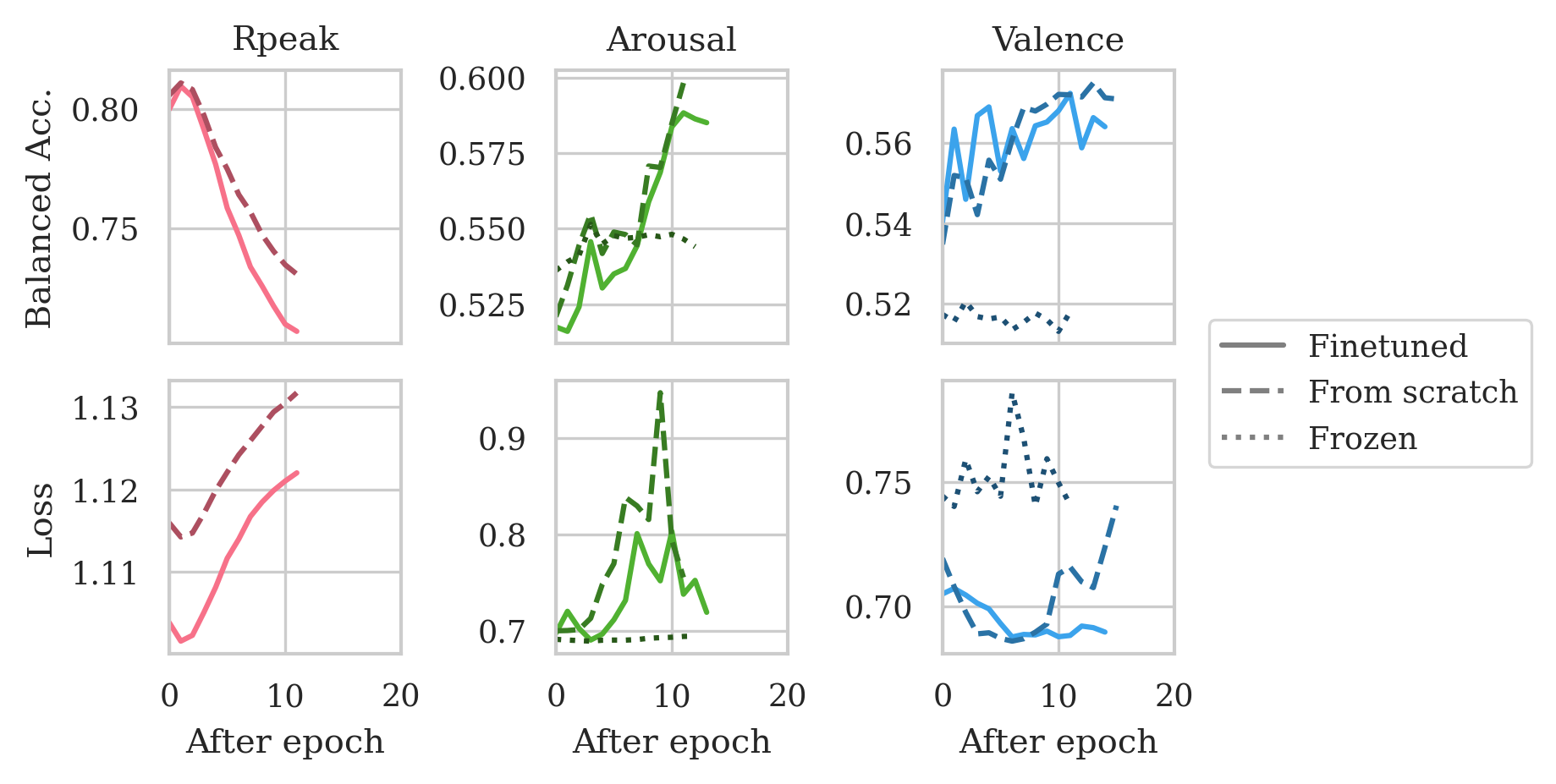}
\caption{Balanced accuracy and loss on the validation set for all model configurations on AffectiveVR, averaged over 5 runs over epochs in which none of the runs was stopped through early stopping on loss.
In arousal and valence, individual runs are not plotted, as they vary too much to be displayed. We do not observe visible differences in convergence speed between the configurations.
}
\label{fig:val_baccs}
\end{figure}
We observe a clear difference between validation and test performance (e.g. validation balanced accuracies of approx. 60\% versus test balanced accuracy of 55.57\% in arousal prediction, see \autoref{tab:crossval_vs_fixed} versus peaks in \autoref{fig:val_baccs}).
High inter-subject variability in affective feedback combined with small sample size in the fixed test and validations sets may lead distortions in performance metrics. Through randomness, particularly easy to classify participants may dominate the validation set, leading to higher performance measures. 
Alternatively the test set may be dominated by hard to classify participants. With 8 subjects, our test split is much smaller than in a common deep learning setting, where test- and validation set are large enough to be considered representative.

To further investigate performance discrepancies, we employ cross-validation for the arousal task. The participants are divided into 5 folds of approximately equal size. The model is trained on 4 folds, with a subset of 4 randomly selected participants from these folds used for validation and early stopping. The remaining fold is used for testing. 
This procedure is repeated 5 times, ensuring that each fold serves as the test set once.
Final performance metrics are calculated by averaging the results across all folds. We use the same hyperparameters as in previous experiments.
This approach, however, includes the validation split, which was used for hyperparameter tuning, in the performance evaluation, introducing a minor information overlap across folds. We assume the impact to be negligibly small. 
\autoref{tab:crossval_vs_fixed} displays performance results on the fixed test split (as reported in \autoref{tab:performance_metrics}) vs results using the described cross-validation. Variance and performance are slightly higher than on the fixed test set, but the \textit{from scratch} configuration remains the best performing.

\begin{table}[tbhp]
\centering
\caption{Cross-validation and fixed-split arousal prediction performance for comparison}
\label{tab:crossval_vs_fixed}
\begin{tabular}{@{}llll@{}}
\toprule
Model       & AUROC                    & B. Acc.                   & F1                       \\ \midrule

\multicolumn{4}{l}{Arousal Prediction (as in \autoref{tab:performance_metrics})}                                                     \\ \midrule
LaBraM ftn. & 0.5621 ± 0.0153          & 54.65\% ± 1.22\%          & 0.5580 ± 0.0391          \\
LaBraM f.s. & \textbf{0.5777 ± 0.0059} & \textbf{55.57\% ± 0.66\%} & \textbf{0.5852 ± 0.0371} \\
LaBraM frz. & 0.5634 ± 0.0036          & 54.48\% ± 0.33\%          & 0.5659 ± 0.0203          \\
CSP-LDA *   & 0.5552 ± 0.0045          & 55.52\% ± 0.51\%          & 0.5498 ± 0.0056          \\ \midrule
\multicolumn{4}{l}{Arousal Prediction (cross-validation on subject level)}                    \\ \midrule
LaBraM ftn. & 0.5721 ± 0.0177          & 54.92\% ± 1.37\%          & 0.5486 ± 0.0409          \\
LaBraM f.s. & \textbf{0.5947 ± 0.0191} & \textbf{55.93\% ± 1.73\%} & \textbf{0.5672 ± 0.0869} \\
LaBraM frz. & 0.5781 ± 0.0191          & 54.91\% ± 1.52\%          & 0.5230 ± 0.1704          \\ \bottomrule  
\end{tabular}
\end{table}

\section{Supplements on attribution patterns}
\label{sec: exhaustive results}
\subsection{Expanded Faithfulness Evaluation}
\label{subsec:faithfulness_eval}

To validate the faithfulness of per-sample Layer-wise Relevance Propagation (LRP) and aggregate heatmaps beyond the results of our R-Peak prediction experiment, we adapt the \ac{MORF}-\ac{AOPC} evaluation proposed by Samek et al. \cite{samekEvaluatingVisualizationWhat2017}. The original method measures how rapidly model outputs degrade toward neutrality as important (i.e., highly scored) features are iteratively removed. The standard MoRF-AOPC formulation is defined as the average drop in the function output over $L$ perturbation steps

\begin{equation}
    \text{AOPC}_{\text{original}} = \frac{1}{L+1} \left\langle \sum_{s=1}^{L} \left( f(x) - f(x_{\text{MoRF}}^{(s)}) \right) \right\rangle
    \label{eq:aopc_original}
\end{equation}

where $\langle \cdot \rangle$ denotes the mean over the test set.

However, Samek et al. conceptualized \ac{AOPC} evaluation for multiclass models, where $f(x)$ is the score of the predicted class and decreases when evidence is removed. Our binary model produces a single logit whose sign is its decision, so removing evidence for a negative decision \emph{increases} $f(x)$. We therefore make two adaptations:

For negative outputs, the summation in Equation~\ref{eq:aopc_original} yields the negative area between the logit and the perturbation curve. To prevent opposite classes from canceling each other out during global averaging, we multiply each sample's sum by the sign of its unperturbed output:

\begin{equation}
    \text{AOPC}_{\text{adapted}} = \frac{1}{L+1} \left\langle \text{sign}(f(x)) \sum_{s=1}^{L} \left( f(x) - f(x_{\text{MoRF}}^{(s)}) \right) \right\rangle
    \label{eq:aopc_adapted}
\end{equation}

We also consider the order in which channels are perturbed, ranking them by their relevance summed over time. For positive decisions ($\mathrm{sign}(f(x)) > 0$), perturbing the most positively relevant channels first should move the logit towards the decision boundary and is thus the appropriate order for an AOPC curve. For negative decisions, the same order would push the logit further into the negative, increasing rather than decreasing the model's confidence. We therefore perturb the most positive channels first where $\mathrm{sign}(f(x)) > 0$ and the most negative channels first where $\mathrm{sign}(f(x)) < 0$, i.e. in descending order of $\mathrm{sign}(f(x)) \cdot \text{relevance}_c$.

To evaluate faithfulness of global attribution maps, as presented in our paper, we establish a single perturbation order based upon them and apply it across the entire test set. Because the heatmaps are aggregated across multiple classification decisions, we determine the perturbation order based on absolute relevance scores, which aligns with our magnitude-based interpretation of heatmaps throughout the main part.
In all perturbation settings, a channel is perturbed by replacing its with the temporal mean value across all channels of the respective sample. Perturbations accumulate across steps. As a baseline, we computed \ac{GXI} \cite{shrikumarNotJustBlack2017} attributions---both per-sample and as a global aggregate---and repeated the identical perturbation procedure.\ac{GXI} attributions are unbounded and can vary strongly between samples, as they do not adhere to a conservation property. To ensure similar contributions of each sample's attribution to the globally ordered AOPC-analysis, we normalize them by their maximum absolute value before aggregation. A per-sample random channel order serves as a chance baseline.

We conduct this evaluation exclusively on the finetuned PhysioMI L/R model from run 3, the best performing run on the established motorimagery/execution task. We employ 10 perturbation steps, in approximate accordance with the amount of relevant electrodes (see \autoref{tab:signif_electrodes}). The results are summarized in \autoref{tab:aopc}. 
\begin{table}[ht]
\centering
\caption{Adapted AOPC (Eq.~\autoref{eq:aopc_adapted}) over 10 perturbation steps, PhysioMI L/R run 3 test split ($n=596$).}
\label{tab:aopc}
\begin{tabular}{lcc}
\toprule
Attribution method & AOPC$_{\text{adapted}}$, mean (SD) & Relative to random \\
\midrule
Per-sample LRP & 0.804 (0.860) & 15.54 \\
Per-sample G$\times$I & 0.340 (0.582) & 6.58 \\
Global aggregate LRP & 0.393 (0.815) & 7.60 \\
Global aggregate G$\times$I & 0.096 (0.322) & 1.86 \\
\midrule
Random baseline & 0.052 (0.271) & 1.00 \\
\bottomrule
\end{tabular}
\end{table}

Per-sample \ac{LRP} displays the highest average \ac{AOPC} by a large margin. In spite of imposing a single perturbation order, the globally aggregated \ac{LRP}-based heatmaps presented in this paper yield a comparable average \ac{AOPC} to per-sample \ac{GXI}, though the global heatmap displays higher variance across samples. Globally aggregated normalized \ac{GXI} heatmaps display only marginally higher faithfulness than the random baseline. Though limited in scope, these results further substantiate our choice of \ac{LRP} as an attribution method for our experiments.

\subsection{Quantitative assessment of electrode relevance consistency}
\label{subsec:quantitative-electrode-consistency}
In the main text, we pointed to some electrodes and patterns as dominant, re-occurringly relevant, or consistent. In an effort to attach quantitative rigor to these qualitative assertions, we conducted statistical analyses on the PhysioMI L/R and arousal state prediction tasks. We compare each electrode ($n = 15$ runs) to the background relevance scores of all other electrodes ($n = (n_{electrodes}-1)*15$. 885 for AffectiveVR, 945 for PhysioMI L/R). One-sided Mann-Whitney U tests are reported (normal approximation with tie correction but without continuity correction). To account for sign-inversion and scale differences between runs, we test on  per-run L1-normalized absolute relevance scores. Benjamini-Hochberg FDR correction is applied across the per-electrode tests. We conduct this testing on the aggregate heatmaps as displayed in our visualizations. Background relevance statistics were virtually identical across electrode tests in both AffectiveVR (Mdn~=~.014, Q1~=~.006, Q3~=~.024--.025 across tests) and PhysioMI (Mdn~=~.012–.013, Q1~=~.006, Q3~=~.021 across tests).
See \autoref{tab:signif_electrodes} for per-electrode test results.

\paragraph{PhysioMI L/R --- assertions substantiated}
Eleven electrodes showed significantly non-random absolute L1-normalized relevance scores: These correspond to the ocular electrodes (AF7, AF8, Fp2, AF3, AF4, Fp1, Fpz) that we point out in the main part. Electrodes from the lateralized pattern described in the main also emerge as significant, together with their opposite-hemisphere counterparts (F7, F5, F8,T9). Additionally, the absolute L1 normalized relevance of the FT7 electrode displays a statistically significant Spearman correlation ($rho= 0.825$, $\text{FDR-adjusted } p = 0.0098$) with generalized test performance. This quantitatively supports our claim that poorly generalizing models rely on ocular shortcuts, while successful ones utilize fronto-temporal lateralized features. 

\begin{table}[ht]
\centering
\small
\setlength{\tabcolsep}{4pt}
\caption{Electrodes with statistical significance.}
\label{tab:signif_electrodes}
\begin{tabular}{llrrrr}
\toprule
El. & Mdn (IQR) & $z$ & $p$ & $p_{\mathrm{FDR}}$ & $r$ \\
\midrule
\multicolumn{6}{l}{\textit{Arousal}} \\
CP2 & .037 (.033--.043) & $5.01$ & $< .001$ & $< .001$ & $.17$ \\
C2 & .043 (.022--.050) & $4.82$ & $< .001$ & $< .001$ & $.16$ \\
Cz & .035 (.026--.037) & $4.60$ & $< .001$ & $< .001$ & $.15$ \\
CP6 & .031 (.024--.039) & $4.19$ & $< .001$ & $< .001$ & $.14$ \\
TP7 & .035 (.024--.040) & $4.17$ & $< .001$ & $< .001$ & $.14$ \\
FC6 & .027 (.023--.043) & $4.16$ & $< .001$ & $< .001$ & $.14$ \\
P6 & .035 (.025--.067) & $3.96$ & $< .001$ & $< .001$ & $.13$ \\
FC2 & .032 (.022--.034) & $3.41$ & $< .001$ & $.002$ & $.11$ \\
POz & .027 (.023--.031) & $3.24$ & $< .001$ & $.004$ & $.11$ \\
TP8 & .026 (.019--.034) & $2.94$ & $.002$ & $.010$ & $.10$ \\
T8 & .025 (.020--.031) & $2.88$ & $.002$ & $.011$ & $.10$ \\
C5 & .026 (.021--.029) & $2.82$ & $.002$ & $.012$ & $.09$ \\
CP5 & .024 (.019--.035) & $2.77$ & $.003$ & $.013$ & $.09$ \\
F3 & .025 (.017--.030) & $2.75$ & $.003$ & $.013$ & $.09$ \\
CPz & .025 (.020--.029) & $2.64$ & $.004$ & $.016$ & $.09$ \\
\midrule \multicolumn{6}{l}{\textit{PhysioMI L/R}} \\
AF7 & .047 (.020--.054) & $4.68$ & $< .001$ & $< .001$ & $.15$ \\
AF8 & .045 (.028--.055) & $4.59$ & $< .001$ & $< .001$ & $.15$ \\
Fp2 & .036 (.024--.050) & $4.25$ & $< .001$ & $< .001$ & $.14$ \\
F7 & .027 (.019--.036) & $4.18$ & $< .001$ & $< .001$ & $.13$ \\
AF3 & .028 (.020--.034) & $4.07$ & $< .001$ & $< .001$ & $.13$ \\
AF4 & .027 (.019--.035) & $3.99$ & $< .001$ & $< .001$ & $.13$ \\
Fp1 & .042 (.025--.052) & $3.80$ & $< .001$ & $< .001$ & $.12$ \\
F5 & .025 (.017--.033) & $3.46$ & $< .001$ & $.002$ & $.11$ \\
F8 & .023 (.020--.028) & $3.14$ & $< .001$ & $.006$ & $.10$ \\
Fpz & .029 (.012--.046) & $2.85$ & $.002$ & $.014$ & $.09$ \\
T9 & .026 (.011--.031) & $2.53$ & $.006$ & $.033$ & $.08$ \\
\bottomrule
\end{tabular}
\end{table}

\paragraph{Arousal state prediction --- confirmation and expansion}
The statistical analysis confirm the significance of the central electrode cluster (C2, Cz, FC2, CP2) and the right parietal-occipital electrode P6 we point out in the main part. Additionally, this approach reveals several additional electrodes with significantly non-random relevance scores:
\begin{itemize}
    \item \textbf{Expanded Central Cluster:} CPz joins the previously established central sensorimotor group.
    \item \textbf{New Temporo-Parietal Cluster:} C5, CP5, and TP7 emerge as an additional, previously unnoticed spatial cluster.
    \item \textbf{P6-Adjacent Cluster:} TP8, T8, and CP6 form a meaningful grouping adjacent to the P6 electrode.
    \item \textbf{Other Emergent Features:} POz, F3, and FC6 also displayed significant relevance.
\end{itemize}
The opposite nature of P6 relevance to that of the central cluster is not verified through these tests and remains a qualitative interpretation.

\subsection{PhysioMI L/R shortcut ablation --- improvement from suspected shortcut electrodes}

\label{subsec:physiomi-ablation}
To establish whether finetuned PhysioMI L/R models indeed relied on facial motorcortex electrodes for generalization, we performed an ablation study by (a) generously excluding suspected left hemisphere facial motorcortex electrodes (T9, T7, FT7, FC5, C5) and a right-hemisphere complement (T10, T8, FT8, FC6, C6), and (b) training exclusively on the aforementioned electrodes. As a control, we repeat training runs with all electrodes. The results are summarized in Table~\ref{tab:mc_ablation}, reporting the mean and standard deviation across 20 fresh runs. The results substantiate our hypothesis: A two-sided Mann-Whitney U test shows that
test balanced accuracy across runs with all available electrodes ($n = 20, Mdn = .736, IQR = .545 - .767$) is significantly higher than test balanced accuracy on runs without facial MC electrodes ($n = 20, Mdn = .530, IQR = .515 - .609$), $U = 281, z = 2.19, p = .028, r = .35$.

\begin{table}[bthp]
\centering

\caption{Performance at varying access to facial motorcortex electrodes. Notation: Mean (std.)}
\label{tab:mc_ablation}
\begin{tabular}{lccc}
\toprule
Condition & AUROC & Bal.\ Acc. & F1 \\
\midrule
All available electrodes & 0.725 (0.149) & 0.664 (0.116) & 0.663 (0.109) \\
Facial MC electrodes only & 0.688 (0.165) & 0.635 (0.120) & 0.622 (0.123) \\
Without facial MC electrodes & 0.612 (0.129) & 0.584 (0.099) & 0.587 (0.100) \\
\bottomrule
\end{tabular}
\end{table}
\subsection{Extended valence attribution pattern analysis}
\label{subsec:extended_valence_attribution_analyisis}
\paragraph{Non-conclusive ambiguity: valence state prediction}
Valence prediction yields a distinct attribution pattern see \autoref{fig:attribution_patterns_dense}, c): 
consistent relevance in the left temporal region (T7, TP7) and, less consistently, in peripheral parieto-occipital electrodes (PO7, O1, O2, PO8). The observed pattern shows only partial overlap with known valence-related EEG findings. Only T7 relevance aligns with Koelstra et al. \cite{koelstraDEAPDatabaseEmotion2012}, while the other regions do not correspond to established correlates. 
Their proximity to face and neck muscles raises the question of \ac{EMG}-artifact-reliance.

\subsection{Step-by-step description of the attribution aggregation process}
\label{subsec:attribution_aggregation_process}
\begin{figure}[tbhp]
    \centering
    \includegraphics[width=\linewidth]{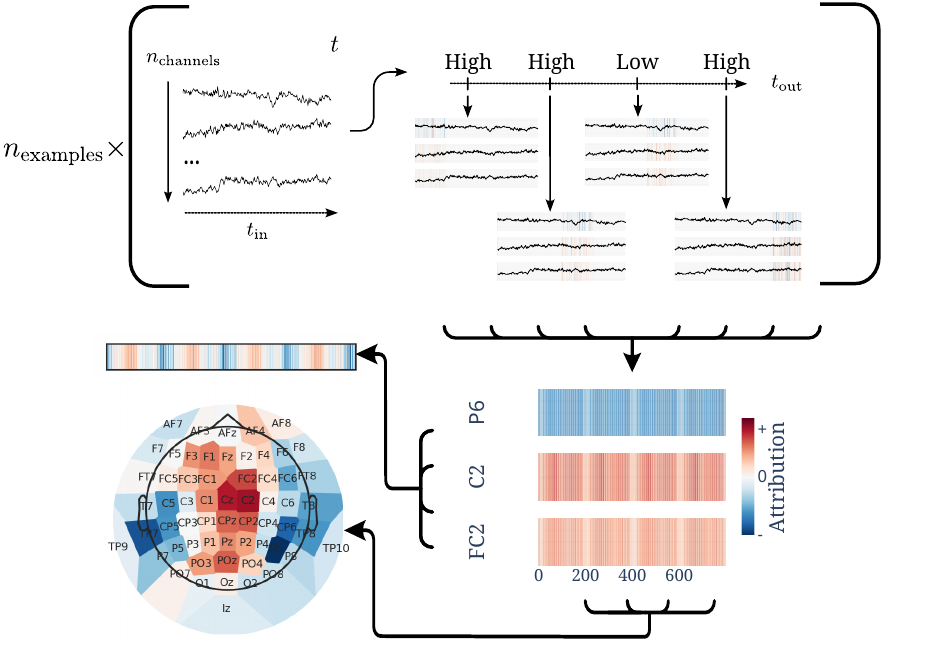}
    \caption{Process for visualizing emerging decision patterns, with dimension annotation. Above: $t_\text{out}$ attributions are created per example, one for each classification decision (here: High or Low arousal/valence). Below: After aggregation over all examples, spatial and temporal aggregates are created. Only three channels displayed for brevity. Joining arrows signify aggregation.}
    \label{fig:attribution_dimensions}
\end{figure}
Following classification, \ac{LRP} is applied to generate attribution scores for each input element.
This is done for each classification decision, that is, for each element in the output.
At an inference step, the model operates on an \ac{EEG} window of $t_{in}$ time steps over $n_{channels}$ each. 
It produces $t_{out}$ logits for prediction of the binarized target. 
Thus, for each \textit{sample} time window, $t_{out} * (t_{in} * n_{channels})$ attribution scores are generated. This is repeated for $n_\text{samples}$ sample time windows.
When $t_{out}$ is large, storage requirements for the analysis quickly become prohibitive. In our R-Peak experiment, we address this by balanced subsampling (see \autoref{sec:data acquisition and cohort details}).
To identify emerging patterns this large set of explained decisions, summarization steps are taken.
We first aggregate attributions across (subsampled) output timesteps of all samples.
This leads to an aggregate in the shape of an input sample ($n_\text{channels} \times t_\text{in}$).
In previous works, rectangular heatmaps are frequently used to visualize attributions of this shape, discarding electrode arrangement.
To facilitate interpretation, we build on the approach from Sturm et al. \cite{sturmInterpretableDeepNeural2016}, which offers temporal and topography-map-like views.
To analyze spatial patterns, we aggregate across time, leading to an aggregate of size $n_\text{channels}$. To analyze temporal patterns, we aggregate across channels, leading to size $t_\text{in}$.
To avoid confusion with activity-related topographic maps, we use discrete Voronoi-cell coloring in our spatial visualizations. \autoref{fig:attribution_dimensions} provides intuition for this attribute-aggregate-visualize approach.

\subsection{Exhaustive per-run attribution patterns}
\label{subsec:exhaustive-per-run-attributions}
\begin{figure}[p]
    \centering
    \includegraphics[width=\textwidth]{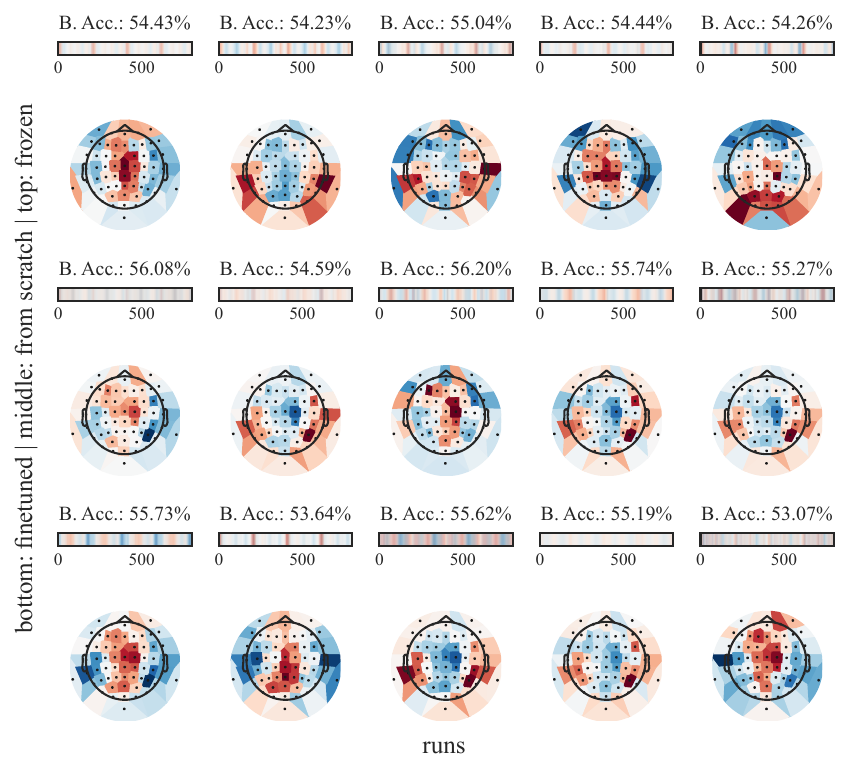}
    \caption{Arousal state prediction patterns on the test split of AffectiveVR, across runs and configurations. Refer to \autoref{paragraph:reporting_and_interpreting_attribution_patterns} for details on the aggregation process.}
    \label{fig:all_arousal_attributions}
\end{figure}

\begin{figure}[p]
    \centering
    \includegraphics[width=\textwidth]{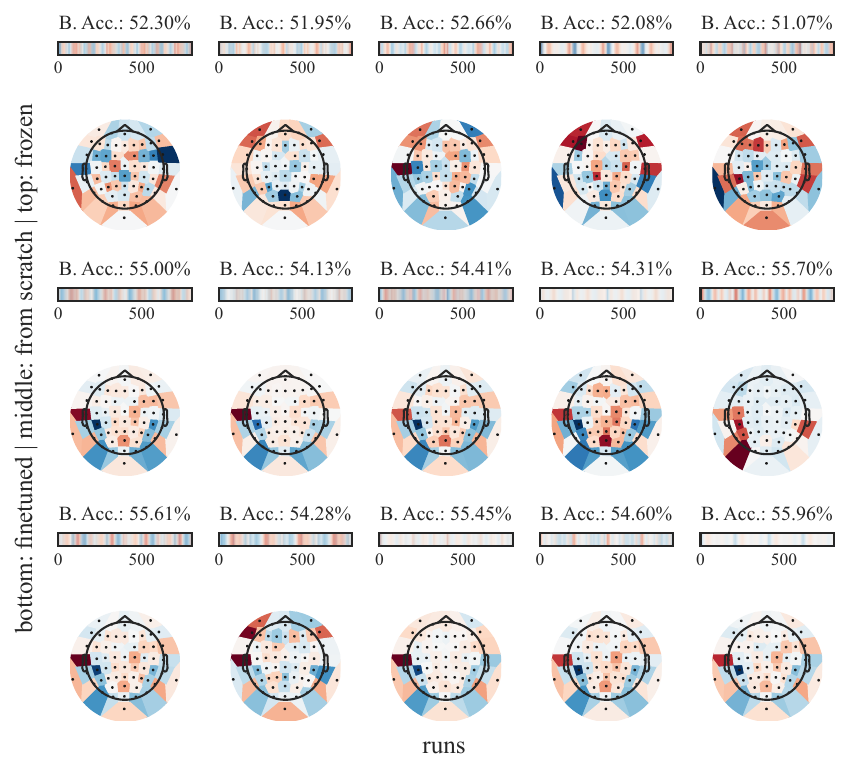}
    \caption{Valence state prediction patterns on the test split of AffectiveVR, across runs and configurations. Refer to \autoref{paragraph:reporting_and_interpreting_attribution_patterns} for details on the aggregation process.}
    \label{fig:all_valence_attributions}
\end{figure}

\begin{figure}[p]
    \centering
    \includegraphics[width=\textwidth]{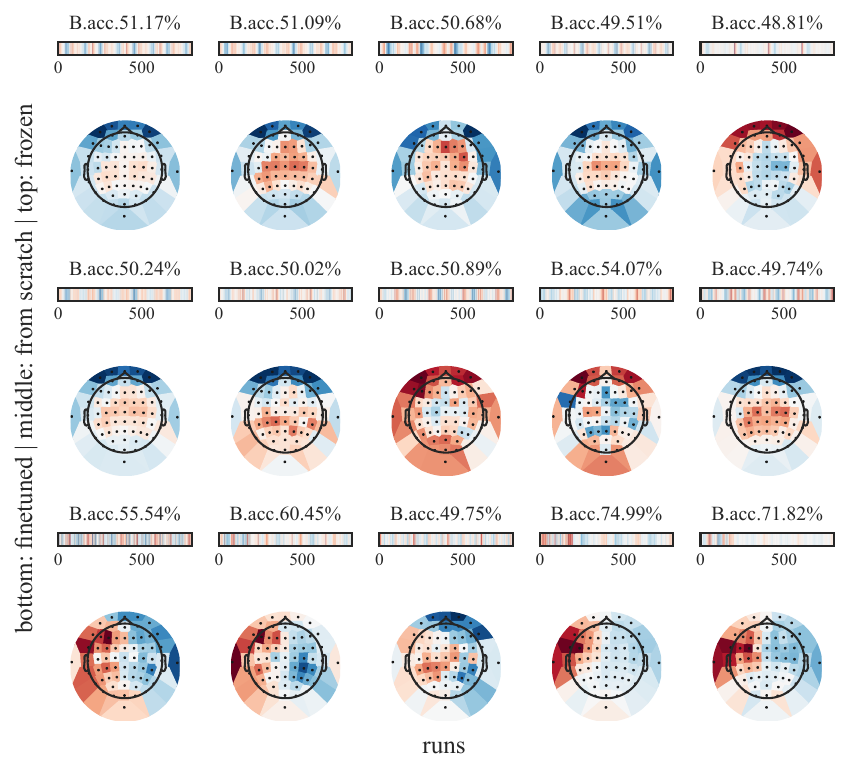}
    \caption{Left/right directional intent prediction patterns on the test split of PhysioMI L/R, across runs and configurations. Refer to \autoref{paragraph:reporting_and_interpreting_attribution_patterns} for details on the aggregation process.}
    \label{fig:all_physiolr_attributions}
\end{figure}

Finally, heatmap-based XAI is visualization intensive. Main-part figures are limited to examples. The attribution patterns for all evaluation runs on their respective test sets are presented in \autoref{fig:all_physiolr_attributions}, \autoref{fig:all_arousal_attributions}, and \autoref{fig:all_valence_attributions} for further consideration. R-Peak prediction patterns are omitted, as they were truly identical across all runs apart from sign flips.
\clearpage
\clearpage
\clearpage
\clearpage

\end{appendices}

\end{document}